\documentclass[letterpaper, 10 pt, conference]{ieeeconf}  

\IEEEoverridecommandlockouts                              

\usepackage{graphics} 
\usepackage{epsfig} 
\usepackage{mathptmx} 
\usepackage{times} 
\usepackage{amsmath} 
\usepackage{amssymb}  

\usepackage{times}
\usepackage[font=footnotesize, skip=10pt]{caption}
\usepackage{multicol}
\usepackage[bookmarks=true]{hyperref}
\usepackage[utf8]{inputenc}
\usepackage{amsmath,amssymb}
\usepackage{algorithm}
\usepackage{algorithmic}
\usepackage{graphicx}
\usepackage{booktabs}
\usepackage{xcolor}
\usepackage{xspace}
\usepackage{verbatim}
\usepackage{siunitx} 
\usepackage{algorithm}
\usepackage{algorithmic}
\usepackage{float} 
\usepackage{pifont}
\usepackage[table]{xcolor}

\let\oldding\ding

\renewcommand{\ding}[1]{%
  \ifnum#1=55
    {\color{red}\oldding{55}}%
  \else
    \oldding{#1}%
  \fi}

\usepackage{multirow}

\newcommand{\method}{OCRA\xspace} 
\usepackage{xcolor}

\newcommand{\fyq}[1]{{#1}}

\author{Kuanning Wang$^{1 *}$, Ke Fan$^{1 *}$, Yuqian Fu$^{1 \dagger}$, Siyu Lin$^{1}$, Hu Luo$^{1}$ \\ Daniel Seita$^{2}$, Yanwei Fu$^{1}$, Yu-Gang Jiang$^{1}$, Xiangyang Xue$^{1 \dagger}$
\thanks{
$^{1}$~Fudan University, China.
$^{2}$~Thomas Lord Department of Computer Science, University of Southern California, USA.
}
\thanks{* indicates equal contributions.}
\thanks{$\dagger$ indicates the corresponding author.}
}

\begin{document}

\title{\LARGE \bf
\method: Object-Centric Learning with 3D and Tactile Priors for Human-to-Robot Action Transfer}

\maketitle
\thispagestyle{empty}
\pagestyle{empty}


\begin{abstract}

We present \method, an \textbf{O}bject-\textbf{C}entric framework for video-based human-to-\textbf{R}obot \textbf{A}ction transfer that learns directly from human demonstration videos to enable robust manipulation. Object-centric learning emphasizes task-relevant objects and their interactions while filtering out irrelevant background, providing a natural and scalable way to teach robots. \method leverages multi-view RGB videos, the state-of-the-art 3D foundation model VGGT, and advanced detection and segmentation models to reconstruct object-centric 3D point clouds, capturing rich interactions between objects. To handle properties not easily perceived by vision alone, we incorporate tactile priors via a large-scale dataset of over one million tactile images. These 3D and tactile priors are fused through a multimodal module (ResFiLM) and fed into a Diffusion Policy to generate robust manipulation actions. Extensive experiments on both vision-only and visuo-tactile tasks show that \method significantly outperforms existing baselines and ablations, demonstrating its effectiveness for learning from human demonstration videos.

\end{abstract}

\section{INTRODUCTION}

``\textit{Man is the most imitative of living creatures, and through imitation learns his earliest lessons}'', observed Aristotle. Humans naturally acquire skills by observing others, motivating the growing paradigm of human-to-robot imitation learning, particularly from videos, which offer a scalable and information-rich source of supervision. Inspired by this, existing work can be broadly categorized into {interface}-based~\cite{Chi2024UniversalMI,liu2024fastumi}, {egocentric vision based}~\cite{nair2022r3m,ma2022vip}, and object-centric based~\cite{Hsu2024SPOTSP}.  Among these, object-centric methods have shown generalization by focusing on the primary objects while reducing sensitivity to less relevant or distracting backgrounds. 
However, approaches such as~\cite{Hsu2024SPOTSP} rely on object-specific priors, such as 6-DOF poses, which limit the range of manipulable objects, and the absence of richer 3D geometry weakens the ability to capture object interactions. 
Other object-centric methods (e.g.,~\cite{Zhu2023LearningGM}) use 3D point clouds, but these are collected through costly, time-consuming teleoperation rather than scalable human videos.

\begin{figure}[t]
  \centering
  \includegraphics[width=1.0\linewidth]{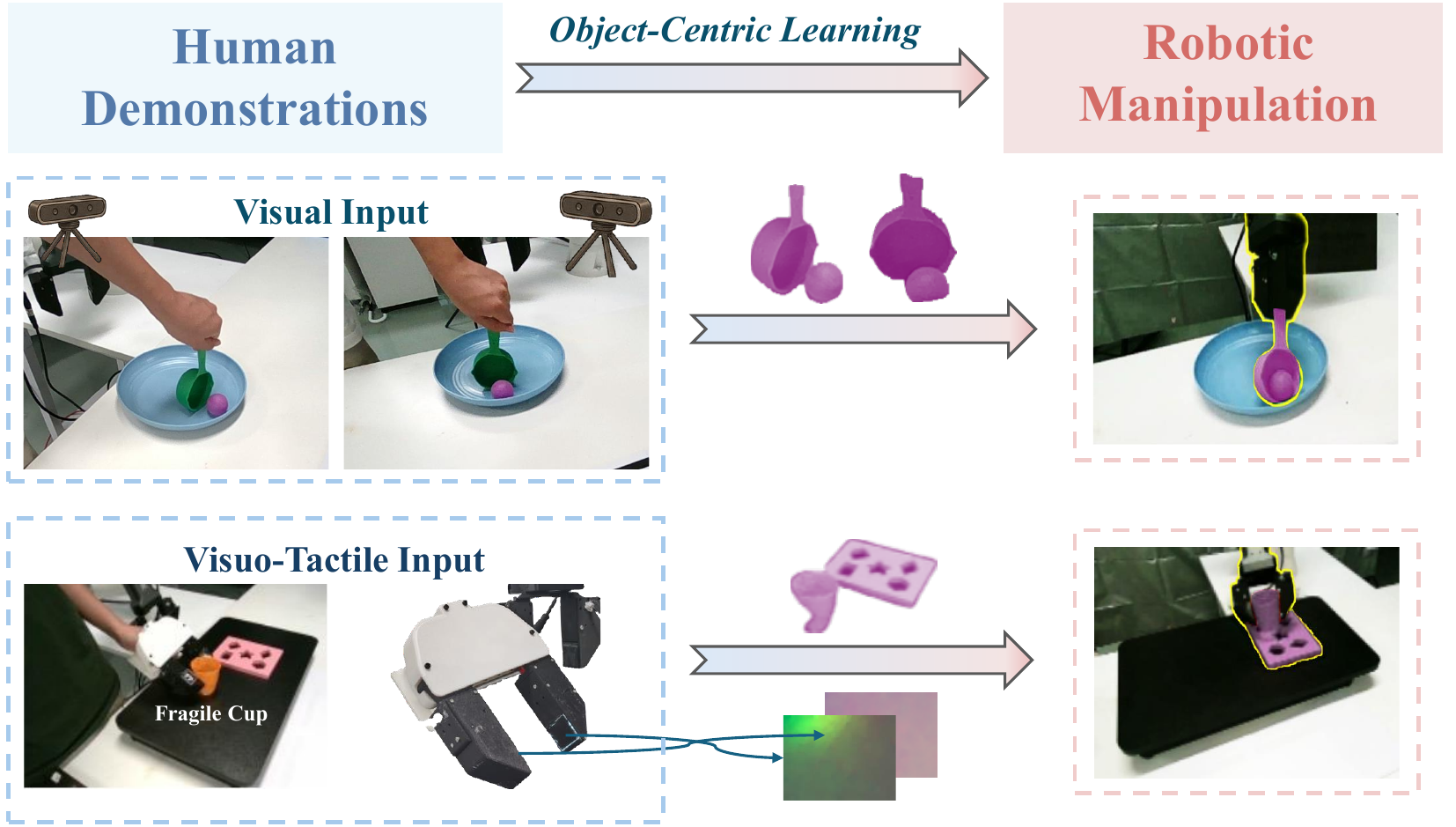}
  \vspace{-0.20in}
  \caption{
  \textbf{Overview of \method}. \method leverages object-centric learning from multi-view human demonstration videos and tactile sensing to enable robust execution of diverse manipulation tasks transferred from humans to robots.
  }
  \label{fig:teaser}
  \vspace{-0.22in}
\end{figure}

In this paper, we propose \method, an \textbf{O}bject-\textbf{C}entric framework for video-based human-to-\textbf{R}obot \textbf{A}ction transfer that combines multi-view RGB inputs, 3D scene reconstruction, and tactile priors to enable robust perception and manipulation. 
Our key contribution is the extraction of rich object-centric representations by integrating 3D information with tactile priors, which are then used to generate effective robot actions directly from human demonstration videos, achieving strong generalization in challenging manipulation tasks.
Specifically, we begin by capturing third-person human manipulation videos. Since single-view RGB often misses critical spatial geometry and interaction details, we record multi-view RGB sequences to provide richer scene coverage. We then leverage VGGT~\cite{Wang2025VGGTVG}, a state-of-the-art 3D foundation model, to reconstruct object-centric 3D scenes and estimate depth, producing more complete and consistent representations than raw depth maps, which are prone to noise, missing regions, and limited coverage. To further isolate task-relevant objects, we apply advanced detection and segmentation models~\cite{Liu2023GroundingDM, Ravi2024SAM2S} to track objects of interest, and combine these object-centric masks with metric depth to extract SE(3) transformations of manipulated objects. By integrating these object-centric visual representations, \method effectively removes background distractions and enables human-to-robot action transfer.

While RGB video provides rich visual information, many manipulation tasks also require tactile perception to capture object properties that vision alone cannot reliably discern. To address this, \method integrates tactile priors collected using a portable tactile gripper, building a dataset of over one million tactile images from diverse object manipulations to pretrain a tactile encoder. This multimodal integration enables the robot to accurately perceive object properties such as texture and weight, allowing robust execution of tasks like sorting objects by their properties and safely adjusting grasp force when handling delicate items.

To fully leverage both visual and tactile information, we present a novel multimodal fusion module, ResFiLM, which integrates 3D and tactile priors to produce rich object-centric representations. These representations are then integrated by our diffusion policy~\cite{Chi2023DiffusionPV} model to generate robust and precise manipulation actions, enabling effective human-to-robot action transfer. 
Extensive experiments show that \method consistently outperforms existing baselines and ablations on both vision-only and visuo-tactile tasks, demonstrating strong performance on challenging manipulation scenarios, thereby validating the effectiveness of our multimodal, object-centric approach.

We summarize our main contributions as, \\
\noindent 1) \textit{Object-Centric Learning from Multi-View Videos}: 
We propose \method, an object-centric framework that focuses on task-relevant objects in multi-view human demonstration videos, using 3D reconstruction and object-centric masks to filter background distractions and capture object interactions for robust human-to-robot action transfer.

\noindent 2) \textit{Integration of 3D and Tactile Priors}: 
We combine 3D object-centric representation produced by 3D foundation model VGGT with tactile priors pretrained from a large-scale tactile dataset via a fusion module (ResFiLM) and a diffusion policy, enabling the robot to perceive object properties and generate precise manipulation actions.

\noindent 3) \textit{Extensive Empirical Validation}: We validate \method across vision-only and visuo-tactile tasks. Our method outperforms baselines and ablations, and the experimental results and analyses highlight its effectiveness and robustness.

\section{RELATED WORK}
\subsection{Imitation Learning from Human Demonstrations Videos}
Traditional imitation learning methods, such as ACT~\cite{zhao2023learning}, Diffusion Policy~\cite{Chi2023DiffusionPV}, and vision-language-action models~\cite{black2024pi_0,kim2024openvla}, primarily focus on behavior cloning~\cite{Pomerleau2015ALVINNAA} from robotic demonstrations. Due to the high cost of collecting such data, recent research increasingly turns to learning from human demonstration videos.
Pretraining approaches like R3M~\cite{nair2022r3m} and VIP~\cite{ma2022vip} address this by leveraging large-scale egocentric human demonstration datasets (e.g., Ego4D~\cite{grauman2022ego4d}) to train visual encoders, which are then fixed for downstream robotic policy learning.
To further narrow the gap between human and robot domains, UMI~\cite{Chi2024UniversalMI} and Fast-UMI~\cite{liu2024fastumi} introduce a universal interface that standardizes the eye-on-hand viewpoint across both human demonstrations and robot executions, enabling more direct policy transfer. However, this restricts demonstrations to the {hand-mounted first-person} perspective.
Alternatively, pose-based methods~\cite{Hsu2024SPOTSP} utilize diffusion models and FoundationPose~\cite{Wen2023FoundationPoseU6} to learn manipulation trajectories from human demonstrations, but they require object-specific priors for accurate pose estimation and additional depth images, while the limited 3D input for policies further hinders their ability to capture complex object interactions.

\subsection{Object-Centric Learning in Robotics}
Object-centric learning~\cite{yuan2023compositional} seeks to represent individual objects rather than entire images by decomposing scenes into semantically coherent regions. Such object-level representations enable robots to perceive and interact with their environments in a manner analogous to human object-centered reasoning. A diverse body of work on object discovery has proposed methods for extracting discrete, meaningful object representations~\cite{jiang2020scalor,linspace,locatello2020object,seitzer2022bridging,kipf2021conditional,singh2022simple}, which have shown effectiveness in reinforcement learning and robotic manipulation~\cite{zadaianchuk2020self,yoon2023investigation}.
More recently, approaches such as FOCUS~\cite{Ferraro2023FOCUSOW} and SOLD~\cite{mosbach2024sold} have embedded these object-centric representations within world models to boost robotic learning. Other lines of research employ explicit object descriptions, such as bounding boxes~\cite{devin2018deep,wang2019deep}, 6-DoF poses~\cite{tremblay2018deep, deng2020self}, or additional annotations, to support downstream robotic tasks. Notably, GROOT~\cite{Zhu2023LearningGM} and VIOLA~\cite{zhu2023viola} construct object representations from segmented or detected instances, thereby enhancing robots’ ability to generalize to novel situations. Progress in this direction has been further accelerated by recent open-world visual foundation models~\cite{Liu2023GroundingDM,kirillov2023segment,Ravi2024SAM2S,caron2021emerging,oquab2023dinov2}.
{Similarly, our work builds on these advances by leveraging visual foundation models to obtain object masks and integrating them with multi-view 3D reconstruction and tactile sensing, yielding richer object-centric representations tailored for human-to-robot action transfer.
}

\subsection{Tactile Sensing Hardware and Learning in Robotics}

Resistive, capacitive, and piezoelectric tactile sensors detect mechanical stimuli via pressure-induced changes in electrical properties. Instead, optical tactile system like GelSight~\cite{yuan2017gelsight} and GelSlim~\cite{donlon2018gelslim} uses a camera to detect soft material deformations, producing optical flow for tactile sensing. 
We use a custom optical tactile sensor inspired by GelSight that generates dense 3D signals via optical flow for normal and tangential forces. Built on open-source code, it is low-cost, user-friendly, and mounted on each finger of a parallel gripper for rich contact sensing. 
3D-ViTac~\cite{Huang20243DViTacLF} combines visual and tactile data to fuse 3D point clouds and uses a diffusion model to learn complex manipulation tasks. It outperforms vision-only methods, even when using low-cost robotic arms. However, it needs teleoperation for the precise states of the robot.
FreeTacMan~\cite{wu2025freetacman} is a wearable, robot-free visuo-tactile demonstration system that records tactile signals along with end-effector poses tracked by a high-precision motion capture system, which limits its applicability in scenarios without such costly tracking infrastructure.

\begin{figure*}[t]
  \centering
  \includegraphics[width=.95\linewidth]{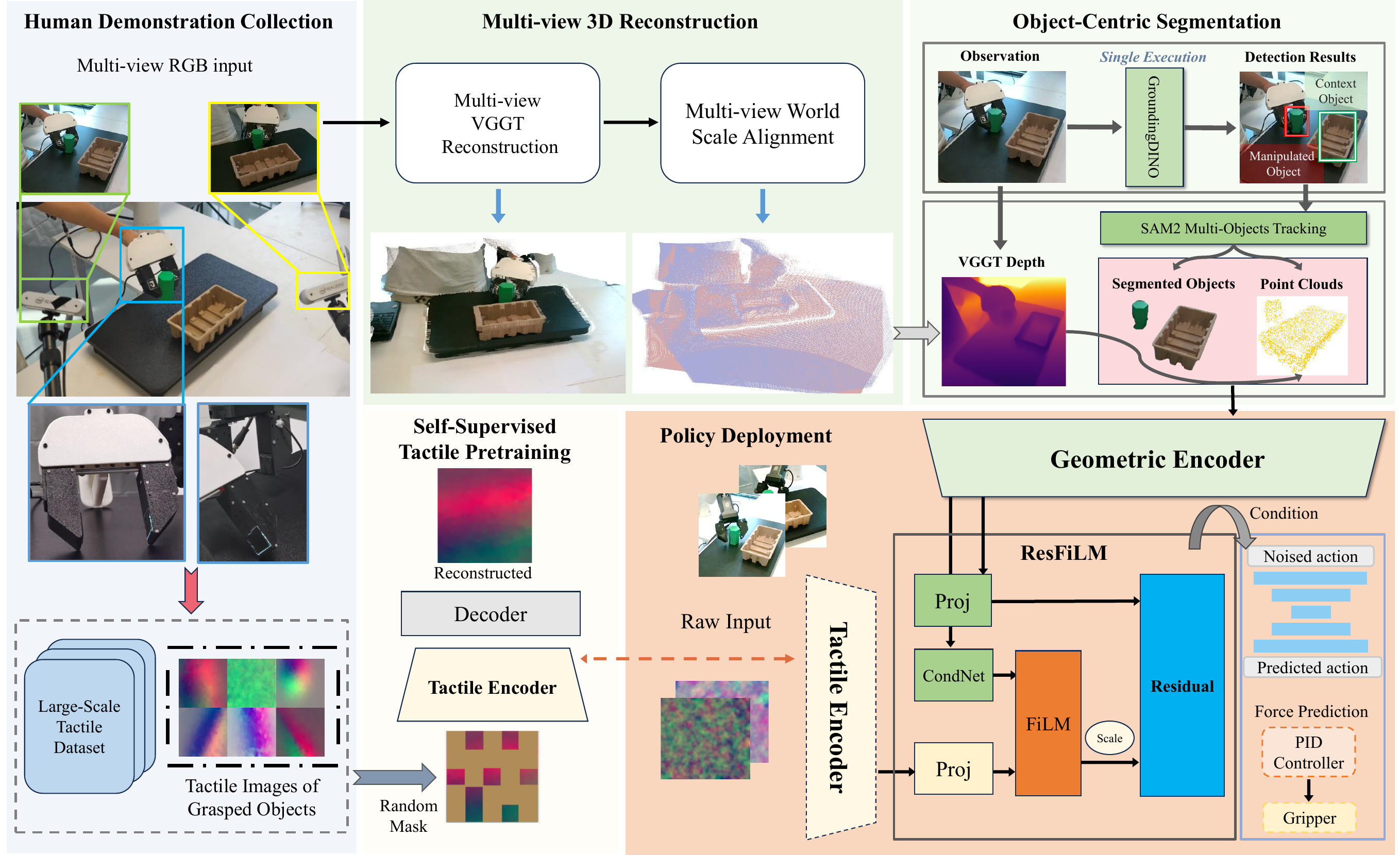}
  \vspace{-0.1in}
  \caption{
    \textbf{Framework.} 
      \underline{\textit{The left column}} illustrates our human demonstration collection system. Two RGB cameras capture demonstration videos, while the blue box highlights a portable tactile gripper for tactile data collection, which also records fingertip tactile images used to build our large-scale tactile dataset (shown at the bottom).
        \underline{\textit{The first row}} depicts how \method processes multi-view RGB inputs to obtain object-centric 3D priors. We first reconstruct the 3D scene using VGGT, followed by bi-view metric depth prediction for world-scale alignment. GroundingDINO and SAM2 then provide object segmentation masks, divided into a Manipulable Object Mask (for target objects) and a Context Object Mask (for surrounding objects). These are used to extract visual object-centric representations across modalities (segmentation, point cloud).
        \underline{\textit{The middle of the second row}} shows tactile-prior extraction via Tactile Encoder pretraining under a Masked Autoencoder paradigm.
        \underline{\textit{The right of the second row}} presents policy deployment. Multi-view RGB and tactile images are encoded into geometric and tactile features, which are fused by ResFiLM and passed to a Diffusion Policy. The policy predicts actions through iterative denoising of noisy action samples.
  }
  \label{fig:framework}
  \vspace{-0.20in}
\end{figure*}

\section{METHOD}
\noindent\textbf{Overview.} 
First, we capture multi-view videos of object manipulation, performed either with the human hand alone or with a gripper controlled by the human. Next, we extract the object-centric 3D prior through 3D reconstruction and object segmentation. The tactile prior is obtained by feature extraction using a large-scale pretrained encoder. Finally, we fuse the visual and tactile signals and use them as conditional inputs to a diffusion model, enabling generative modeling of the object trajectory.
\subsection{Human Demonstration Collection System}

\noindent\textbf{Hardware Design.}
We designed an easy-to-use system to collect human demonstrations, as shown in the left column of Fig.~\ref{fig:framework}. 
For the visual information, we used 2 external cameras to capture RGB information from a third-person perspective during human demonstrations of object manipulation. The task-relevant objects were simultaneously captured by the two cameras, whose relative positions were calibrated.

\noindent\textbf{Vision-Based Tactile Sensor.}
To capture tactile information, we employ an elastomeric sensing layer embedded with tracer particles, providing
a compact, low-cost design with high-resolution sensing capability,  following~\cite{luo2024novel}. Under external loading, the particles exhibit motion patterns governed by the Poisson effect, which vary with the applied force. By analyzing the resulting displacement field within the material, we extract tactile signals. 
To ensure stability and reduce errors from ambient light, we add a black light-shielding mask and controlled LED lighting.

\subsection{Object-Centric 3D Prior}

\noindent\textbf{Multi-view 3D Reconstruction.} 
We used multiple RGB perspectives; however, simply using RGB perspectives overlooks the relationship between images. To make more effective use of this 3D prior, we used a 3D reconstruction model to extract the scene point cloud.
We adopt the state-of-the-art multi-view reconstruction algorithm VGGT~\cite{Wang2025VGGTVG}, which estimates scene geometry from multi-view RGB images and performs real-time processing of RGB frames to produce depth maps, point maps, and pixel-wise tracking. This approach not only reduces the negative impact of sensor noise but also makes the method more practical, since RGB images are much easier to obtain and more widely available than RGB-D data.

Considering the scale ambiguity in the point maps by VGGT, we measure the transformation matrix between the two camera views and feed it into the depth prediction branch. By comparing the relative camera poses predicted by VGGT with the measured camera poses, we calibrate the depth prediction and achieve metric depth estimation. 

\noindent\textbf{Object-Centric Point Cloud Segmentation.}
To obtain object point clouds, we leverage advanced vision foundation models. 
Specifically, a text description is first provided to GroundingDINO~\cite{Liu2023GroundingDM} to generate target bounding boxes in the initial frame. These bounding boxes then serve as prompts for SAM2~\cite{Ravi2024SAM2S}, enabling real-time segmentation and tracking of the targets throughout the two third-person videos $V^{{view},1}$ and $V^{{view},2}$.
We segment all task-relevant objects in the video, dividing the resulting masks into the manipulated object mask $M^{{manip}}$ and the context object mask $M^{{ctx}}$. Since two cameras are used, we perform segmentation separately on each view, crop the corresponding depth regions, and back-project them into 3D space according to their respective camera poses. The resulting point clouds from both views are then fused, yielding the manipulated object point cloud $P^{{manip}}$ and the context object point cloud $P^{{ctx}}$. The whole process is illustrated in Fig.~\ref{fig:framework}.

\noindent\textbf{Object Point Cloud Encoder.} 
We extract features by concatenating the object-centric point clouds $P^{manip}/P^{ctx}$ with $M^{manip}/M^{ctx}$, which indicate manipulated and context objects, as input to the Geometric Encoder, following DP3~\cite{Ze20243DDP}.

\noindent\textbf{Object Transformation Estimation.} 
To obtain \textit{pose transformation} of the the manipulate object between adjacent frames, we apply the Iterative Closest Point (ICP) algorithm to compute the rotation and translation $T_t\in SE(3)$ at each time step $t$, aligning the manipulated object point clouds $P_t^{{manip}}$ and $P_{t+1}^{{manip}}$ computed from $V^{view}$.
Benefiting from multi-view reconstruction, the object point clouds provide rich geometry, enabling reliable and effective ICP-based transformation estimation across frames.
\subsection{Tactile Prior}
\noindent\textbf{Flow-based Tactile.} 
The \textit{displacement field} of the sensor is computed by comparing the optical flow between the first frame \(V^{tac}_{0}\) and the current frame \(V^{tac}_{t}\). The first frame is initialized under unloaded conditions, while the current frame may be either loaded or unloaded. We use the Dense Inverse Search (DIS) algorithm to estimate the optical flow. 
The optical flow is further decoupled into 3-D force vectors \(F^{tac}_{t}\in\mathbb{R}^{240\times320\times3}\) with a U-Net, following~\cite{luo2024novel}.

\noindent\textbf{Large-Scale Tactile Pretraining.} 
We use our portable gripper to collect a large-scale dataset of \underline{one million tactile} images, covering diverse objects, gripping forces, gripping postures, and manipulation tasks. Among them, a large subset consists of high-quality samples with coarse annotations of the manipulated objects and their corresponding task labels.
We adapt a ViT-based Tactile Encoder to extract features from tactile images. To mitigate the domain gap, we do not use pretrained weights from natural images; instead, we pretrain the encoder on our self-collected tactile dataset. The encoder is trained in a self-supervised manner using the Masked Autoencode~\cite{He2021MaskedAA} paradigm.

\subsection{Generative Trajectory Diffusion Modeling}
\noindent\textbf{Diffusion Framework.}
We model the distribution of future object transformations using a generative model over $SE(3)$ transformations. This approach captures complex, multimodal dynamics conditioned on past observations, and outperforms direct per-step regression methods.
Recent Diffusion Policies sample joint-space action chunks via Langevin dynamics; to decouple from robot configuration, we instead predict the manipulated object’s rigid-body motion conditioned on past observations.
\begin{equation}
\label{eq:diffusion_policy_ours}
    {T}^{k-1}_t = \alpha_k({T}^k_t - \gamma_k\epsilon_\theta({O}_t,{T}^k_t,k) + \sigma_k\epsilon^k),
\end{equation}
where $T_t^k\in\mathbb{R}^{T \times 4\times4}$, $\epsilon^k$ is a Gaussian noise, and $\alpha_k,\gamma_k,\sigma_k$ are parameters of reverse process.
The corresponding loss is defined as:
\begin{equation}
    \mathcal{L} = \mathbb{E}_{k,\epsilon^k} \|{\epsilon}^k -  \epsilon_\theta(T_t^0+{\epsilon}^k,k,O_t)\|_2^2.
    \label{eq:diffusion_policy_ours_loss}
\end{equation}

\noindent\textbf{Conditioning Network.}
Learning from human demonstration videos is hindered by the embodiment gap, the discrepancy between human demonstrators during training and robotic agents during deployment.
To overcome this challenge, we utilize a fused object-centric feature representation $\mathbf{f}_{oc}$, which isolates and emphasizes the object of interest to ensure consistency across both phases.

To integrate tactile features into the object-centric point cloud feature, we design a lightweight \textbf{ResFiLM} (\textbf{Res}idual \textbf{FiLM}) fusion module. Given a point-cloud feature $\mathbf{f}_{pc} \in \mathbb{R}^{D}$ and aggregated tactile features $\mathbf{f}_{t} \in \mathbb{R}^{D}$, the module first projects them into a common latent space. Conditioned on the visual feature, we generate FiLM parameters $(\gamma, \beta)$ that modulate the tactile feature, yielding a tactile embedding $\mathbf{f}_{t}' = \gamma \cdot \mathbf{f}_{t} + \beta$. The final fusion is performed via a gated residual connection:
\begin{equation}
\mathbf{f}_{oc} = \mathbf{f}_{pc} + \alpha\,(\boldsymbol{\gamma}\odot\mathbf{f}_{t}+\boldsymbol{\beta}),
\end{equation}
where $\alpha$ is a learnable scaling coefficient. This design ensures stable training and allows tactile signals to act as adaptive residual corrections to the visual representation.

\subsection{Robot Control}

We use the external camera's extrinsic parameter to transfer the motion of the object in the camera's frame to the robot's frame. We transfer the SE(3) transformations of the object to the ones of the robot gripper. 
At inference step $t$, given the sampled transformation $T_t^0$ (we omit the diffusion step $0$ hereafter), the object is moved by the cumulative transformation
$\prod_{i=1}^{t} T_i = T_t \cdots T_1.$

In addition to the predicted transformation, we augment the action representation with tactile feedback by appending the average contact force from the left and right finger pads. This enables the policy to explicitly utilize force information during control, rather than relying solely on geometric motion cues.
The gripper is controlled with a PID-based scheme that takes the contact force predicted from demonstrations (rescaled for calibration) as the reference. Tactile feedback is incorporated during grasping to maintain a stable hold, enabling reliable force-controlled manipulation.

\begin{figure*}[t]
  \centering
  \includegraphics[width=1.\linewidth]{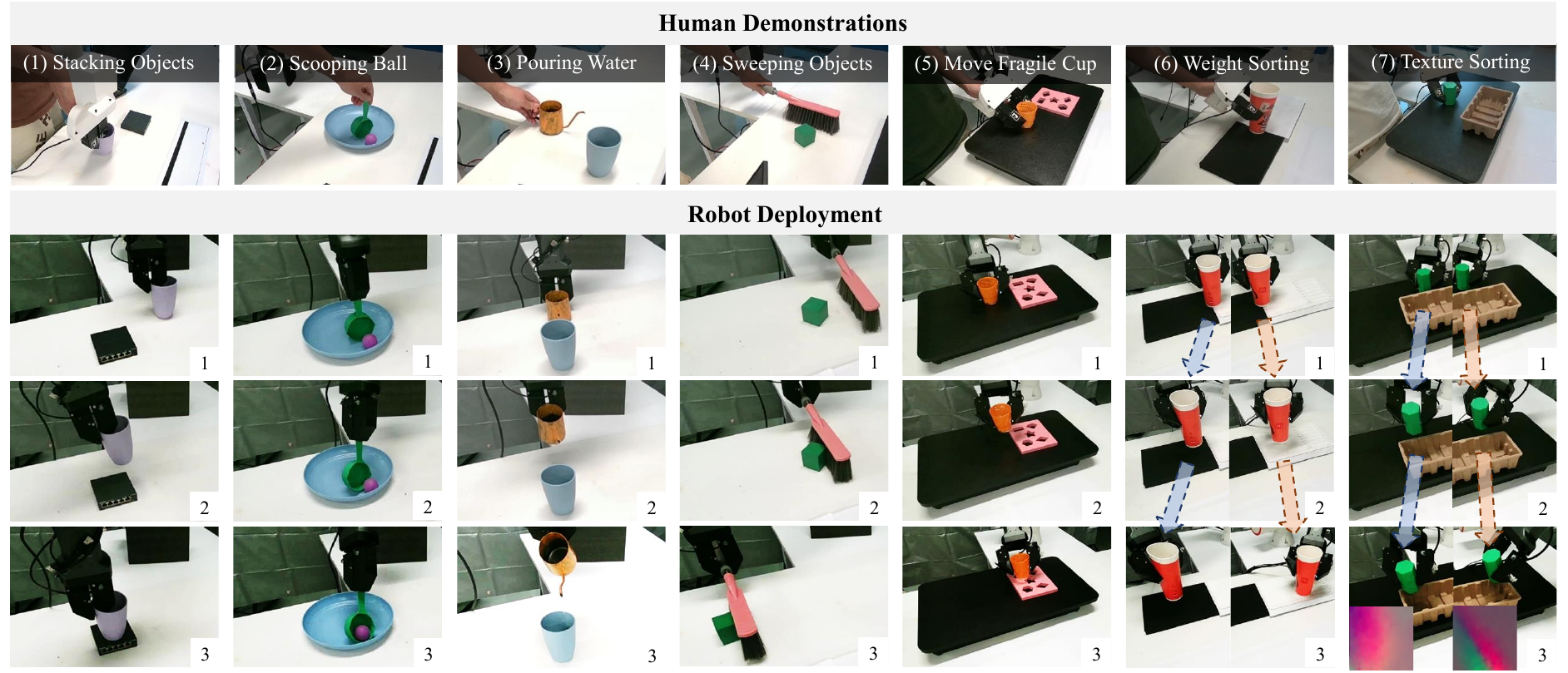}
  \vspace{-0.25in}
  \caption{
      \textbf{Experiments Visualization}. The first row shows a human demonstrator using either a hand or a portable gripper to collect demonstrations. The left four tasks are vision-only, while the right three are visuo-tactile. The following rows show the robot’s execution of our policy.
      The last two columns illustrate visuo-tactile tasks under different input conditions: in the Robot Deployment subfigures, the left and right insets correspond to different execution attempts, with arrows of different colors indicating motion directions. In the Weight Sorting task, cups are guided to distinct target locations based on mass; in the Texture Sorting task, objects are sorted by surface texture. For the Texture Sorting column (column 7, inset 3), we additionally show tactile images from grasps on different textures. These results demonstrate that vision alone is insufficient for reliable discrimination, highlighting the critical role of tactile perception in accurate decision-making.
  }
  \label{fig:comb_exp}
  \vspace{-12pt}
\end{figure*}

\begin{figure}[t]
  \centering
  \includegraphics[width=1.\linewidth]{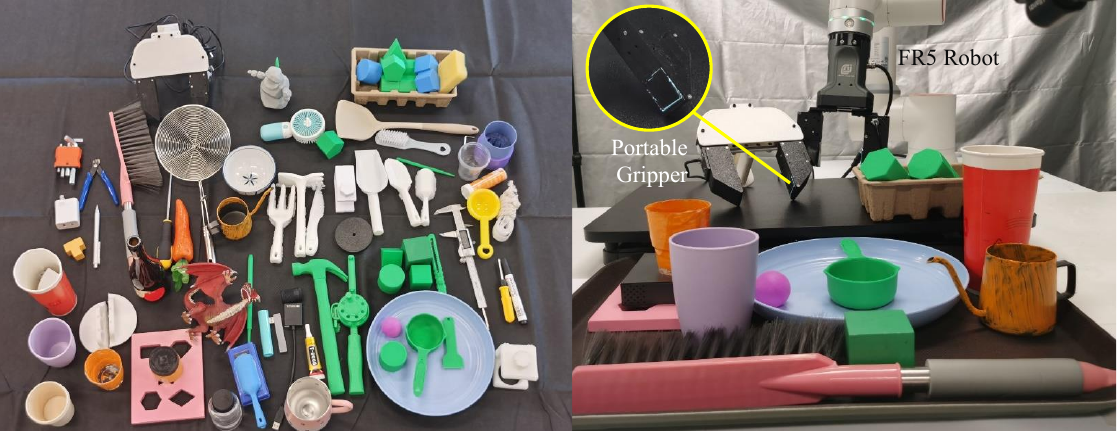}
  \caption{
      \textbf{Experimental setup for our system}.  
      For the right part, the top half shows the robot and portable gripper, which uses the same tactile device. The bottom half shows our experimental objects. We display the camera setting in Fig.~\ref{fig:framework}.
      For the left part, the image shows a subset of the objects we used to collect our large-scale tactile image dataset.
  }
  \label{fig:setup}
  \vspace{-16pt}
\end{figure}

\section{EXPERIMENTS}

\subsection{Experimental Setup}
\noindent\textbf{Hardware.} A workstation with an NVIDIA RTX 6000 Ada GPU is used for policy training and evaluation. A low-cost FAIRINO Robot 5 and Jodell Gripper are used for real-world manipulation. We have an NVIDIA A800 GPU for tactile encoder pretraining. 
We use 2 RealSense-D455 as the visual cameras.
The human-held portable gripper is 3D-printed.

\noindent\textbf{Implementation.}
Depending on task complexity, we collected 20 to 40 human demonstrations for each task, ensuring sufficient diversity in both demonstrations and evaluations (Time cost in Tab.~\ref{tab:demo_time}).
 Our policy is trained by learning rate $1e-4$, with 
 the observation horizon  2, and  predicted steps  4 or 8, decided by the task.
We begin each trial by moving the robot and gripper to a predefined pre-grasp pose.

\noindent\textbf{Evaluation Protocols.}
Evaluation on 7 tasks—4 vision-only and 3 requiring tactile input. Each task runs 10 trials; we report human-judged mean success, discarding trials with external disturbances or hardware failures, so rates reflect intrinsic policy skill in precise, contact-rich manipulation:

\noindent \textbf{{Vision-Only} Tasks}: (1) \textit{{Stacking Objects.}} This task requires the robot to grasp a plastic cup and place it on a box.
{A trial succeeds if the cup stays on the box’s top surface.}
(2) \textit{{Scooping Ball.}} This task requires the robot to use a spoon to scoop a ball from a plate.
{Success requires lifting the ball and keeping it in the spoon.}
(3) \textit{{Pouring Water.}} This task requires the robot to manipulate a teapot and attempt to pour water into a nearby cup. The task is considered successful if water enters the cup. 
(4) \textit{{Sweeping Objects.}} This task requires the robot to use a broom to sweep a block across the table surface. {Moving the block in the intended direction with a steady broom motion is regarded as a success.}

\noindent \textbf{Visuo-Tactile Tasks}: (5) \textit{{Move Fragile Cup.}}
{This task requires the robot to grasp and move a lightweight, fragile cup to a target place (a pink box), testing its ability to manipulate delicate objects. A trial is successful if the cup is placed without falling or noticeable deformation (e.g., crushing or bending).}
(6) \textit{{Weight Sorting.}} {This task requires the robot to recognize a given cup's weight (empty or weighted). Placing the empty cups in the white area {or} the weighted cups in the black area is regarded as a success. }
(7) \textit{Texture Sorting.} {The robot distinguishes and sorts two visually similar objects with different geometric textures (hexagonal vs. octagonal prisms), placing each on its designated side of the tray. A trial is successful if both objects are correctly positioned.}

\subsection{Baselines}
\noindent For those vision-only tasks, we evaluate our method without tactile information, where all tactile-related modules are removed and the policy is conditioned solely on 3D point clouds. We consider two baseline variants:
(a) \textbf{RGB-based Diffusion Policy}, abbreviated as ``DP (RGB)": The policy is trained conditioned only on RGB images {without 3D point clouds}. 
{The object trajectories are still used as supervision during training to ensure consistent guidance across methods.}
(b) \textbf{3D-based Diffusion Policy without Object-Centric Modeling}, abbreviated as ``DP (w/o OC)": We extract 3D scene point clouds from VGGT, and feed them directly into the policy through our point cloud encoder, without applying object-centric modeling.

{For those visuo-tactile tasks, we compare our complete \method against ``\textbf{Ours without Tactile}", where all tactile-related modules are discarded, and the policy relies solely on point cloud observations.}

\section{Results}

\subsection{Vision-Only Results}
We show our Vision-Only results in Tab.~\ref{tab:realexp}.
Our \method method achieves an 85.0\% success rate across four challenging manipulation tasks. These tasks span a diverse range of difficulties: fluid dynamics in \textit{Pouring Water}, dynamic object handling and tool–object interactions in \textit{Scooping Ball} and \textit{Sweeping Objects}, and precise object manipulation in \textit{Stacking Cup}. The results highlight that our approach can robustly address tasks requiring reasoning over both rigid and non-rigid object behaviors.
Our qualitative results could be seen in Fig.~\ref{fig:comb_exp}. 
Notably, in the highly challenging \textit{Scooping Ball} scenario, we further evaluated the case where the ball is placed at the center of the bowl, a configuration that induces unstable dynamics. Remarkably, we observed that \method could robustly follow the ball’s motion through repeated collisions and still accomplish a successful scoop.

\begin{table}[t]
  \centering
  \small
  \scalebox{0.9}{
  \begin{tabular}{l|l|c|c|c} 
  \toprule
  \rowcolor{blue!5}
  \textbf{Task} & \textbf{Method} & \textbf{Success} & \textbf{Proc. Fail.} & \textbf{Out. Fail.} \\
  \midrule
  \multirow{3}{*}{Stack} 
    & Ours        & 100\% & 0\% & 0\% \\
    & DP (RGB)    & 0\%  & 100\% & 0\% \\
    & DP (w/o OC) & 10\%  & 50\%  & 40\% \\
  \midrule
  \multirow{3}{*}{Scoop} 
    & Ours        & 70\%  & 30\% & 0\% \\
    & DP (RGB)    & 0\%  & 70\% & 30\% \\
    & DP (w/o OC) & 0\%  & 100\% & 0\% \\
  \midrule
  \multirow{3}{*}{Pour} 
    & Ours        & 70\%  & 0\% & 30\% \\
    & DP (RGB)    & 0\%  & 70\% & 30\% \\
    & DP (w/o OC) & 30\%  & 20\% & 50\% \\
  \midrule
  \multirow{3}{*}{Sweep} 
    & Ours        & 100\% & 0\% & 0\% \\
    & DP (RGB)    & 90\%  & 0\% & 10\% \\
    & DP (w/o OC) & 50\%  & 50\% & 0\% \\
  \bottomrule
  \end{tabular}
  }
  \vspace{-0.08in}
  \caption{\textbf{Real-world manipulation task evaluation.}  
Outcomes are categorized into Success, Process Failure (Proc. Fail.), and Outcome Failure (Out. Fail.).}
  \label{tab:realexp}
  \vspace{-0.25in}
\end{table}

We further compare against baseline methods. The RGB-based Diffusion Policy achieves partial success; for instance, it can sweep objects, but often fails in other tasks. 
Typical errors include cups being lifted but moved randomly in stacking, and the spoon scooping prematurely and pushing the ball away.
In pouring tasks, misalignment between the spout and the container is common. Overall, the policy struggles to generalize from limited demonstrations, with performance further degraded by variations in background, lighting, and manipulator differences (human versus robot). \fyq{This indicates the benefit of incorporating 3D information.}

Similar to the RGB-based variant, the Diffusion Policy without object-centric modeling achieves a substantially lower success rate than \method. While 3D spatial information offers modest gains in complex tasks such as Pouring Water and Stacking Objects, generalization remains limited due to point cloud noise and train–test discrepancies. In Sweeping Objects, performance is even worse than the RGB baseline, often stalling mid-execution. \fyq{This supports our motivation for adopting an object-centric learning approach.}

\textbf{\fyq{More Analysis of the Performance Gap.}} 
\fyq{To further investigate the relatively low performance of the two baselines, we examined their failure cases and categorized them into two types:}
\textit{outcome failures} and \textit{process failures}. Outcome failures occur when a full trajectory is executed but ends inaccurately. Process failures arise when the robot cannot complete a coherent trajectory. The two baselines predominantly suffer from process failures, indicating an inability to generate or sustain coherent trajectories, while our method eliminates such failures, \fyq{demonstrating robustness in producing executable trajectories.
}

In summary, these comparison studies demonstrate that incorporating an object-centric 3D representation is critical for enabling robust generalization across diverse manipulation tasks. Our \method significantly outperforms baselines by leveraging spatial information and object centric understanding, thereby achieving state-of-the-art success rates in both rigid and non-rigid object interactions.

\begin{table}[t]
  \centering
  \small
  \scalebox{.8}{
  \begin{tabular}{l|c|c|c|c} 
  \toprule
  \rowcolor{blue!5}
  \textbf{Method} & \textbf{Fragile Cup} & \textbf{Weight Sort} & \textbf{Texture Sort}  & \textbf{Average} \\
  \midrule
  Ours      & 90\%  & 90\% & 90\%  & 90\% \\
  Ours(W/o Tactile) & 60\% & 30\% & 30\% & 40\% \\
  \bottomrule
  \end{tabular}
  }
  \vspace{-0.08in}
  \caption{ 
  Success rates of different methods on three evaluated real-world manipulation tasks
  }
  \label{tab:realexp2}
  \vspace{-0.1in}
\end{table}

\subsection{Visuo-Tactile Results}
Main visuo-tactile results are shown in Tab.~\ref{tab:realexp2}, where our \method with tactile information achieves a 90\% success rate. 
See the qualitative results in Fig.~\ref{fig:comb_exp}.
{RGB-based and 3D-based DP fail on Fragile Cup and Texture Sort, with less than 30\% success rate on Weight Sort, so we omit them in the table.}
Specifically, in the \textit{Move Fragile Cup} task, our \method enables the gripper to securely grasp the cup with slight, non-damaging deformation, allowing stable lifting and placement without breakage in all trials; in the \textit{Weight Classification} task, cups are consistently grasped without damage. Heavy cups are placed on the right and empty cups on the left, supported by tactile shear force signals that provide discriminative weight information~\cite{RoboticCompliant}; in the \textit{Texture Sort} task, the two prismatic objects are visually similar, making vision alone insufficient. Tactile sensing enabled reliable category discrimination by guiding placement decisions, while vision planned the trajectory.

Besides, for \method without tactile feedback, the average success rate dropped to 40\%. Vision alone could not distinguish subtle shape differences or detect cup weight, causing frequent errors in the sorting tasks. In the fragile cup task, inconsistent grasping forces led to cup deformation, unnatural trajectories, and placement failures.

\begin{table}[t]
  \centering
  \small
  \scalebox{.9}{
  \begin{tabular}{l|c|c|c} 
  \toprule
  \rowcolor{blue!5}
  \textbf{Method} & \textbf{Success} & \textbf{Proc. Fail.} & \textbf{Out. Fail.} \\
  \midrule
    \textbf{Ours} & 90\% & 0\% & 10\% \\
    Ours (w/o Pretrain) & 30\% & 20\% & 50\% \\
    Ours (w/o Fusion) & 30\% & 40\% & 30\% \\
  \bottomrule
  \end{tabular}
  }
  \vspace{-0.08in}
  \caption{ 
  \textbf{Ablations on the \textbf{Texture Sort} task.}  
  Outcomes categorized as Success, Process Failure (Proc. Fail.), and Outcome Failure (Out. Fail.).
  }
  \vspace{-0.25in}
  \label{tab:ablation}
\end{table}

\subsection{Ablations}
After establishing the vision-only baseline, we further investigate how tactile information contributes to policy learning in the visuo-tactile setting. To this end, we conduct two ablation studies: (a) \textbf{Without tactile pretraining}: The tactile encoder is randomly initialized, while the normalization statistics remain derived from the large-scale dataset; (b) \textbf{Without Residual FiLM Fusion}: The fusion module is removed, and point cloud features are directly concatenated with tactile features.
As shown in Tab.~\ref{tab:ablation}.

Our analysis indicates that our proposed masked tactile pre-training enhances the feature discriminability of tactile inputs, enabling the model to extract more informative and noise-resilient features, thereby improving the stability of both trajectory generation and final stage execution. Furthermore, 
the design of the fusion module is critical for generating reliable trajectories. When point cloud and tactile features are simply concatenated, the model is easily misled by noisy tactile signals, resulting in erroneous and unstable trajectories, proving the necessity of our Residual FiLM. {Similar phenomenon is observed in Weight Sort Task.}

\begin{figure}[t]
  \centering
  \includegraphics[width=.95\linewidth]{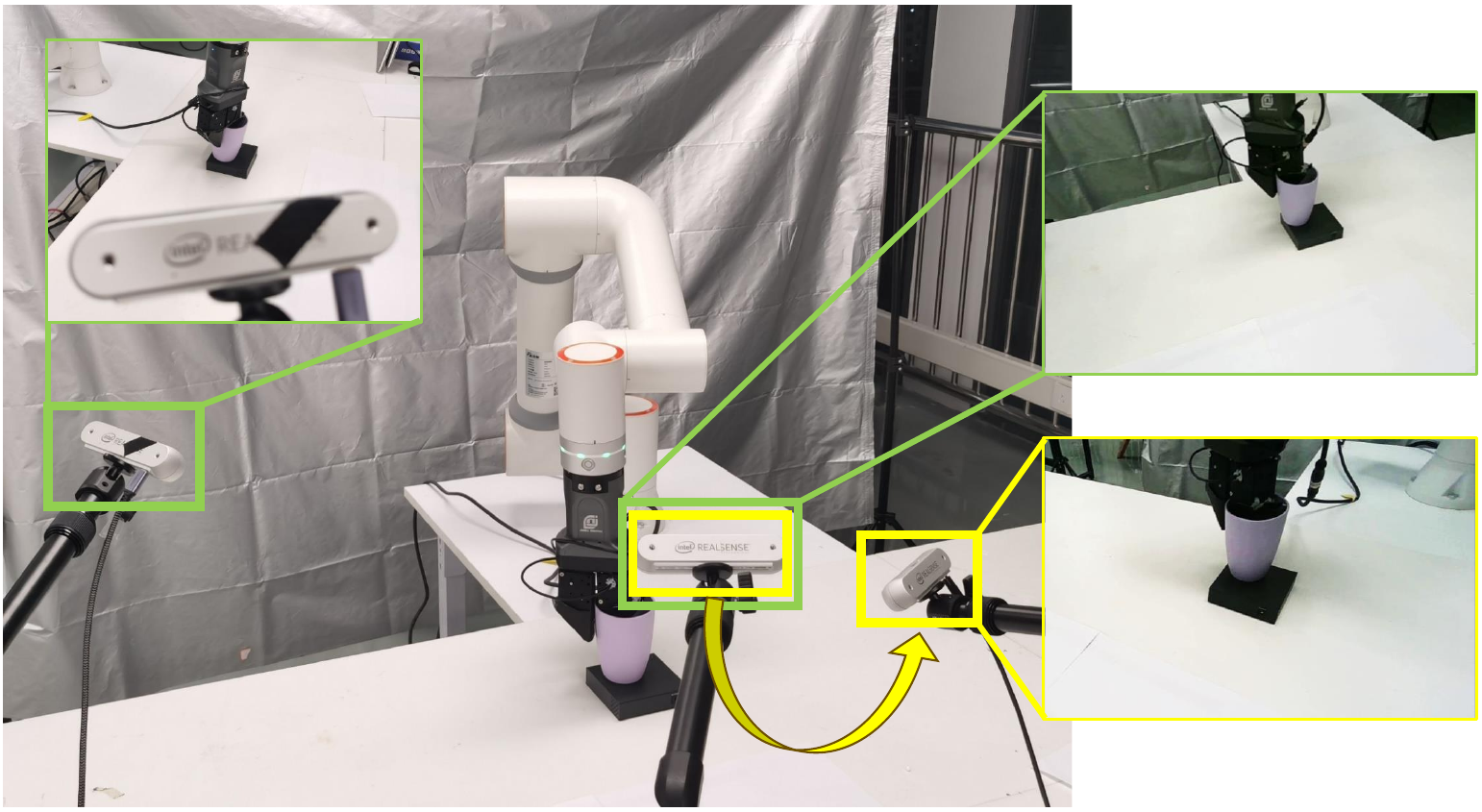}
  \vspace{-0.1in}
  \caption{
      \textbf{Visualization of camera views}. We use the two cameras marked by green boxes during both the demonstration collection and testing. To evaluate the view generalization ability of \method, the left green-boxed camera remains fixed, while the camera in both green and yellow boxes is replaced according to the arrow direction. 
  }
  \label{fig:view_general}
  \vspace{-0.12in}
\end{figure}

\subsection{View Generalization}
A common challenge in visuomotor learning is sensitivity to camera viewpoints, where even small perspective changes can affect perception and action execution. To evaluate whether our method overcomes this challenge, we replaced one of the original cameras with a novel third-view camera and tested it on the \textit{Stacking Objects} task, which critically depends on recognizing object spatial relationships. 
In 10 out of 10 trials, our method successfully completed the task with smooth trajectories.
Remarkably, performance remained stable even when the cup is partially outside the new camera’s field of view, indicating that our approach reuses learned perceptual representations in a viewpoint-invariant manner rather than memorizing specific camera configurations. Qualitative results in Fig.~\ref{fig:view_general} further illustrate robust generalization under unseen viewpoints.

\begin{table}[t]
  \centering
  \small
  \scalebox{.9}{
  \begin{tabular}{l|c|c|c|c|c} 
  \toprule
  \rowcolor{blue!5}
  \textbf{Method} & \textbf{Stack} & \textbf{Sweep} & \textbf{Scoop} & \textbf{Pour} & \textbf{Tactile Tasks} \\
  \midrule
  \textbf{Ours} & 8 & 8 & 8 & 8 & 5 \\
  \textbf{Teleoperation} & 20 & 20 & N/A & N/A & N/A \\
  \bottomrule
  \end{tabular}
  }
  \caption{\textbf{Demonstration collection time (seconds per task).} 
``N/A'' denotes tasks that are not applicable or cannot be performed by the method. 
Our approach achieves substantially lower demonstration time and greater adaptability compared to teleoperation.}
\vspace{-0.2in}
  \label{tab:demo_time}
\end{table}

\begin{figure}[t]
  \centering
  \includegraphics[width=1.\linewidth]{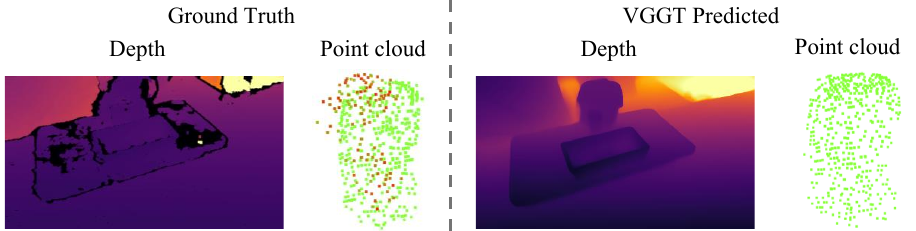}
  \vspace{-0.12in}
  \caption{
      \textbf{Visualization of depth reconstruction quality.} We compare the depth map generated by VGGT with the camera ground-truth (GT) depths from the RGBD cameras on the Texture Sort task. The GT depth contains numerous holes and often produces misaligned point clouds due to measurement instability (see red points in the figure), whereas VGGT yields smoother depth maps and more consistent, regular point clouds.
  }
  \label{fig:reconst_dep}
  \vspace{-10pt}
\end{figure}

\begin{table}[t]
  \centering
  \small
  \scalebox{.8}{
  \begin{tabular}{l|c|c} 
  \toprule
  \rowcolor{blue!5}
  \textbf{Task} & \textbf{Depth Setting} & \textbf{Success Rate} \\
  \midrule
  Stack & Ours & 100\% \\
        & Single-View GT Depth & 60\% \\
  \midrule
  Texture Sort & Ours & 90\% \\
               & Bi-view GT Depth & 50\% \\
  \bottomrule
  \end{tabular}
  }
  \caption{\textbf{Depth analysis across tasks}. Success rates for our reconstruction compared with the camera's ground-truth (GT) depth baselines. }
  \vspace{-0.15in}
  \label{tab:depth_ablation}
\end{table}

\subsection{Further Analysis}

\noindent\textbf{Demonstration Collection Efficiency.}
Tab.~\ref{tab:demo_time} compares demonstration times across tasks. Our method consistently requires only 8 seconds per demonstration (5 seconds for tactile tasks), achieving over a 2$\times$ speedup compared to teleoperation (20 seconds). Moreover, teleoperation is infeasible for \textit{Scoop} and \textit{Pour} due to complex object–object dynamics, and for tactile tasks due to the lack of direct force feedback. In contrast, our approach handles all tasks efficiently, highlighting both its speed and adaptability.

\noindent\textbf{Flexibility to Ground-Truth Depth During Inference.}  
Our framework is designed as a depth-free approach, leveraging multi-view RGB reconstruction for both data processing and deployment. Nevertheless, when ground-truth depth is available (e.g., from depth sensors), it can be directly used during testing {with mild performance drop (less than 20\%) on Stacking, Sweeping and Texture/Weight Sort.} 

\noindent\textbf{Comparing \method with Ground-Truth Depth.}
Tab.~\ref{tab:depth_ablation} presents a comparison between our \method with 3D reconstruction and OCRA models trained and tested using ground-truth depth baselines.
On the \textit{Stack} task, which relies solely on vision, single-view depth fails to capture object–object interactions, leading to collisions. On the \textit{Texture Sort} task, which combines vision and touch, bi-view depth yields only 50\% success, as small objects suffer from missing regions in the depth maps and merging views introduces misalignment from calibration errors. 
As shown in Fig.~\ref{fig:reconst_dep}, ground-truth depth often suffers from holes and noisy point clouds, while VGGT yields smoother and more consistent geometry. This richer reconstruction enables more reliable execution, explaining the performance gap in Tab.~\ref{tab:depth_ablation}.

\section{CONCLUSIONS}
\fyq{
We presented \method, an object-centric framework for human-to-robot imitation learning directly from RGB demonstration videos. By integrating multi-view reconstruction with VGGT, object-centric point cloud segmentation and feature extraction, large-scale tactile data collection and pretraining, and generative trajectory diffusion modeling with a lightweight multimodal (3D+Tactile) fusion module, \method provides rich object-centric representations that enable robust action transfer from human to robot. Experiments on both vision-only and visuo-tactile tasks validate the effectiveness of this design, consistently outperforming baselines. 

}

\section*{ACKNOWLEDGMENT}

Prof. Yanwei Fu is the core project leader.  The paper is supported by   Science and Technology   Commission of Shanghai Municipality (No. 24511103100), and National Natural Science Foundation of China (62521004).




\bibliographystyle{IEEEtran}
\bibliography{main}

\end{document}